\documentclass[preprint,12pt]{elsarticle}
\usepackage{epsfig}
\usepackage{amssymb}
\usepackage{amsmath}
\usepackage{amsthm}
\usepackage{caption}
\usepackage{subcaption}
\usepackage{mathabx}

\newcommand{\ignore}[1]{}

\usepackage{lipsum}
\makeatletter
\def\ps@pprintTitle{%
 \let\@oddhead\@empty
 \let\@evenhead\@empty
 \def\@oddfoot{}%
 \let \@oddfoot}
\makeatother
\begin{document}

\begin{frontmatter}

\title{Understanding Convolutional Neural Networks with A Mathematical Model}

\author{C.-C. Jay Kuo}
\address{Ming-Hsieh Department of Electrical Engineering \\
University of Southern California, Los Angeles, CA 90089-2564, USA}

\begin{abstract}

This work attempts to address two fundamental questions about the
structure of the convolutional neural networks (CNN): 1) why a
nonlinear activation function is essential at the filter output of all
intermediate layers? 2) what is the advantage of the two-layer cascade
system over the one-layer system?  A mathematical model called the
``REctified-COrrelations on a Sphere" (RECOS) is proposed to answer
these two questions. After the CNN training process, the converged
filter weights define a set of anchor vectors in the RECOS model. Anchor
vectors represent the frequently occurring patterns (or the spectral
components). The necessity of rectification is explained using the RECOS
model.  Then, the behavior of a two-layer RECOS system is analyzed and
compared with its one-layer counterpart.  The LeNet-5 and the MNIST
dataset are used to illustrate discussion points. Finally, the RECOS
model is generalized to a multilayer system with the AlexNet as an
example. 

\begin{keyword}
Convolutional Neural Network (CNN), Nonlinear Activation, RECOS Model,
Rectified Linear Unit (ReLU), MNIST Dataset.
\end{keyword}

\end{abstract}

\end{frontmatter}

\section{Introduction}

There is a strong resurging interest in the neural-network-based
learning because of its superior performance in many speech and
image/video understanding applications nowadays. The recent success of
deep neural networks (DNN) \cite{juang2016deep} is due to the
availability of a large amount labeled training data (e.g. the ImageNet)
and more efficient computing hardware. It is called deep learning since
we often observe performance improvement when adding more layers.  The
resulting networks and extracted features are called deep networks and
deep features, respectively.  There are two common neural network
architectures: the convolutional neural networks (CNNs)
\cite{Nature2015} and the recurrent neural networks (RNNs).  CNNs are
used to recognize visual patterns directly from pixel images with
variability. RNNs are designed to recognize patterns in time series
composed by symbols or audio/speech waveforms.  Both CNNs and RNNs are
special types of multilayer neural networks.  They are trained with the
back-propagation algorithm. We will focus on CNNs in this work. 

Although deep learning tends to outperform classical pattern recognition
methods experimentally, its superior performance is somehow difficult to
explain. Without a good understanding of deep learning, we can only have
a set of empirical rules and intuitions. There has been a large amount
of recent efforts devoted to the understanding of CNNs.  Examples
include scattering networks
\cite{mallat2012group,bruna2013invariant,wiatowski2015mathematical},
tensor analysis \cite{cohen2015expressive}, generative modeling
\cite{dai2014generative}, relevance propagation \cite{bach2015pixel},
Taylor decomposition \cite{montavon2015explaining}, etc.  Another
popular topic along this line is on the visualization of filter
responses at various layers 
\cite{simonyan2013deep,zeiler2014visualizing,zhou2014object}. 

It is worthwhile to point out that the CNN is a special form of the
feedforward neural network (FNN), also known as the multi-layer
perceptron (MLP), trained with back-propagation. It was proved in
\cite{hornik1989multilayer} that FNNs are capable of approximating any
measurable function to any desired accuracy. In short, FNNs are
universal approximators. The success of CNNs in various applications
today is a reflection of the universal approximation capability of FNNs.
Despite this theoretical foundation, the internal operational mechanism
of CNNs remains mysterious. 

This research attempts to address two fundamental questions about CNNs:
1) Why a nonlinear activation operation is needed at the filter output
of all intermediate layers?  2) What is the advantage of the cascade of
two layers in comparison with a single layer? These two questions are
related to each other.  The convolutional operation is a linear one. If
the nonlinear operation between every two convolutional layers is
removed, the cascade of two linear systems is equivalent to a single
linear system. Then, we can simply go with one linear system and the
necessity of a multi-layer network architecture is not obvious. Although
one may argue that a multi-layer network has a multi-resolution
representation capability, this is a well known fact and has been
extensively studied before.  Examples include the Gaussian and the
wavelet pyramids. There must be something deeper than the
multi-resolution property in the CNN architecture due to the adoption of
the nonlinear activation unit. 

The existence of nonlinear activation makes the analysis of CNNs
challenging.  To tackle this problem, we propose a mathematical model to
understand the behavior of CNNs.  We view a CNN as a network formed by
basic operational units that conducts ``REctified COrrelations on a
Sphere (RECOS)". Thus, it is called the RECOS model. A set of anchor
vectors is selected for each RECOS model to capture and represent
frequently occurring patterns. For an input vector, we compute its
correlation with each anchor vector to measure their similarity.  All
negative correlations are rectified to zero in the RECOS model, and the
necessity of rectification is explained.  

Anchor vectors are called filter weights in the CNN literature. In the
network training, weights are first initialized and then adjusted by
backpropagation to minimize a cost function.  Here, we adopt a different
name to emphasize its role in representing clustered input data in the
RECOS model.  After the analysis of nonlinear activation, we examine
two-layer neural networks, where the first layer consists of either one
or multiple RECOS units while the second layer contains only one RECOS.
We conduct a mathematical analysis on the behavior of the cascaded RECOS
systems so as to shed light on the advantage of deeper networks.  The
study concludes by analyzing the AlexNet which is an exemplary
multi-layer CNN. 

\begin{figure}
\centering
\includegraphics[width=12cm]{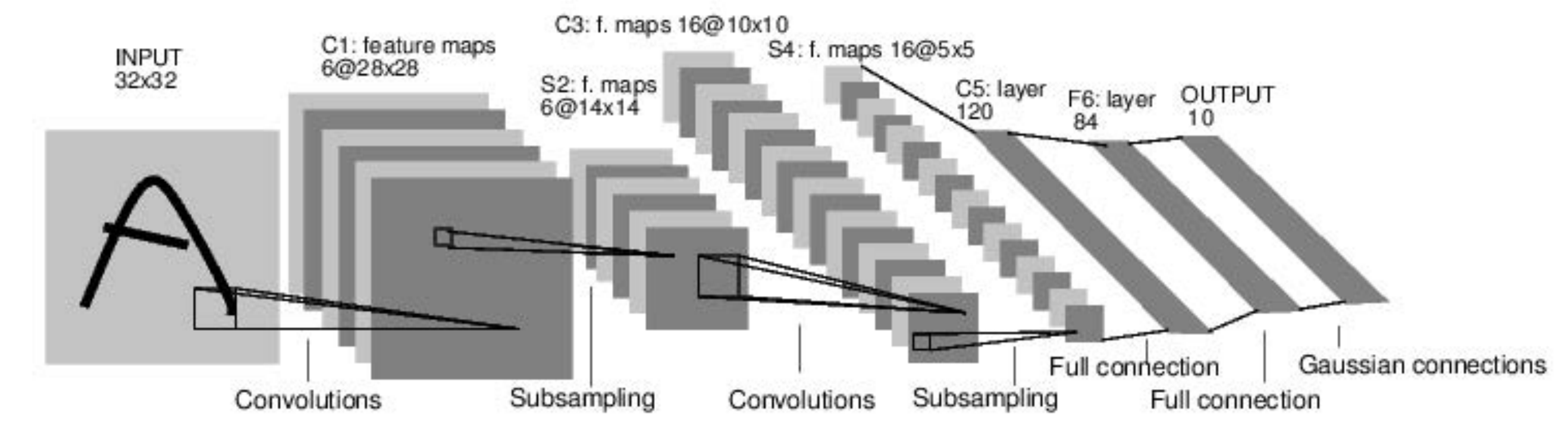}
\caption{The LeNet-5 architecture \cite{LeNet1998}.}\label{fig:LeNet}
\end{figure}

To illustrate several discussion points, we use the LeNet-5
applied to the MNIST dataset as an example. The MNIST
dataset\footnote{http://yann.lecun.com/exdb/mnist/} is formed by ten
handwritten digits (0, 1, ..., 9).  All digits are size-normalized and
centered in an image of size 32 by 32.  The dataset has a training set
of 60,000 samples and a test set of 10,000 samples.  The LeNet-5 is
the latest CNN designed by LeCun et al. \cite{LeNet1998} for handwritten
and machine-printed character recognition. Its architecture is shown in
Fig.  \ref{fig:LeNet}. The input image is an 8-bit image of size 32 by
32. The LeNet-5 has two pairs of convolutional/pooling layers, denoted
by C1/S2 and C3/S4 in the figure, respectively.  C1 has 6 filters of
size 5 by 5.  C3 has 16 filters of size 5 by 5. Each of them is followed
by a nonlinear activation function (e.g. the sigmoid function).
Furthermore, there are two fully connected layers, denoted by C5 and F6,
after the two pairs of cascaded convolutional/pooling/clipping
operations and before the output layer.  The LeNet-5 has a strong impact
on the design of deeper networks in recent years. For example, the
AlexNet proposed by Krizhevsky {\em et al.} in \cite{NIPS2012_AlexNet}
is a generalization of the LeNet-5 from two compound
convolutional/pooling/activation layers to five. 

\section{Why Nonlinear Activation?}

Generally speaking, CNNs attempt to learn the relationship between the
input and the output and store the learned experience in their filter
weights. One challenge to understand CNNs is the role played by the
nonlinear activation unit after the convolutional operation.  We will
drop the pooling operation in the discussion below since it mainly
provides a spatial dimension reduction technique and its role is not as
critical. 

\begin{figure}
\centering
\includegraphics[width=12cm]{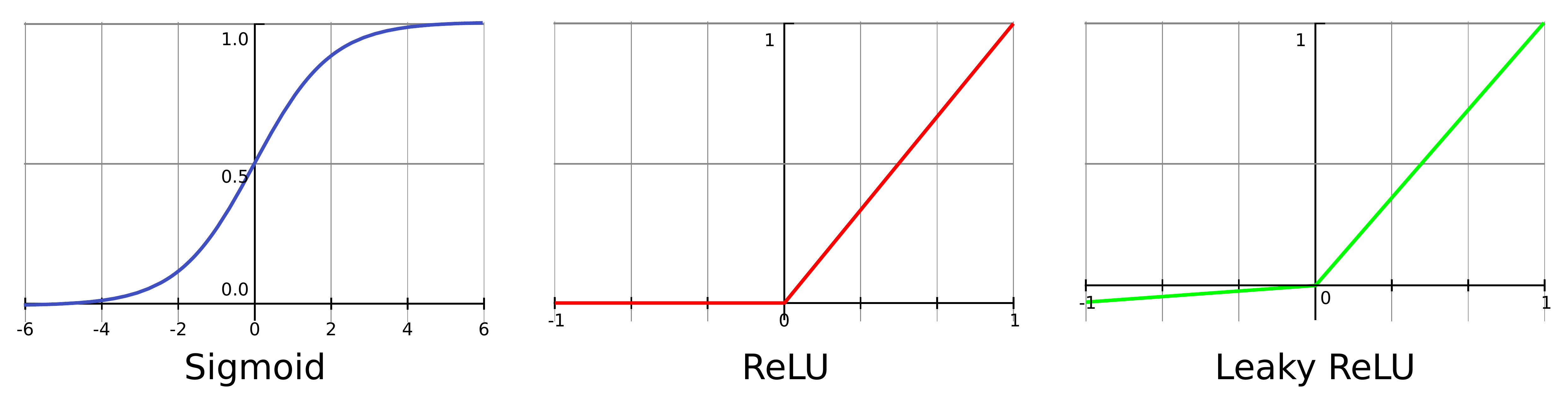}
\caption{Three nonlinear activation functions adopted by CNNs: the sigmoid
function (left), the ReLU (middle) and the Leaky ReLU (right).}\label{fig:clipping}
\end{figure}

The adoption of nonlinear activation in neural networks can be dated
back to the early work of McCulloch and Pitts
\cite{mcculloch1943logical}, where the output of the nonlinear
activation function is set to $1$ or $-1$ if the input value is positive
or non-positive, respectively. A geometrical interpretation of the
McCulloch-Pitts neural model was given in \cite{zhang1999geometrical}. 

In the recent literature, three activation functions are commonly used
by CNNs.  They are the sigmoid function, the rectified linear unit
(ReLU) and the parameterized ReLU (PReLU) as shown in Fig.
\ref{fig:clipping}. The PReLU is also known as the leaky ReLU.  All of
them play a clipping-like operation.  The sigmoid clips the input into
an interval between 0 and 1. The ReLU clips negative values to zero
while keeping positve values unchanged.  The leaky ReLU has a role
similar to the ReLU but it maps larger negative values to smaller ones
by reducing the slope of the mapping function.  It is observed
experimentally that, if the nonlinear operation is removed, the system
performance drops by a large margin. 

Each convolutional layer is specified by its filter weights which are
determined in the training stage by an iterative update process.  That
is, they are first initialized and then adjusted by backpropagation to
minimize a cost function.  All weights are then fixed in the testing
stage.  These weights play the role of ``system memory''.  In this work,
we adopt a different name for filter weights to emphasize their role in
the testing stage.  We call them ``anchor vectors'' since they serve as
reference signals (or visual patterns) for each input patch of test
images.  It is well known that signal convolution can also be viewed as
signal correlation or projection.  For an input image patch, we compute
its correlation with each anchor vector to measure their similarity.
Clearly, the projection onto a set of anchor vectors offers a spectral
decomposition of an input. 

Anchor vectors are usually not orthogonal and under-complete. Consider
the LeNet-5. For the first convolutional layer (C1), the input patch is
of size $5 \times 5=25$. It has 6 filters (or anchor vectors) of the
same size. Thus, the dimension and the number of anchor vectors at C1
are 25 and 6, respectively.  For the second convolutional layer (C3),
its input is a hybrid spatial/spectral representation of dimension $(5
\times 5) \times 6=150$. Then, the dimension and the number of anchor
vectors in C3 are 150 and 16, respectively.

Here, we interpret the compound operation of ``convolution followed by
nonliear activation" as a mechanism to conduct ``REctified COrrelations
on a Sphere (RECOS)". Without loss of generality, we adopt the ReLU
activation function in the following discussion.  All negative
correlations are rectified to zero by the ReLU operation in the RECOS
model. The necessity of rectification is explained below.  To begin
with, we consider a unit sphere centered at the origin. 

{\bf Origin-Centered Unit Sphere.} Let ${\bf x}=(x_1, \cdots, x_N)^T$
be an arbitrary vector on a unit sphere centered at the origin in
the $N$-dimensional space, denoted by
\small
\begin{equation}\label{eq:unitsphere}
S=\left\{ {\bf x} \biggr\rvert ||{\bf x}||=(\sum_{n=1}^N x^2_n)^{1/2}=1 \right\}.
\end{equation}
\normalsize
We are interested in clustering ${\bf x}$'s with the geodesic
distances over $S$. The geodesic distance between vectors ${\bf x}_i$
and ${\bf x}_j$ in $S$ is proportional to the magnitude of their angle,
which can be computed by
\begin{equation}\label{eq:angle}
\theta ({\bf x}_i, {\bf x}_j) = \cos^{-1} ({\bf x}^T_i {\bf x}_j).
\end{equation}
Since $\cos \theta$ is a monotonically decreasing function for $0^o \le
|\theta| \le 90^o$, we can use the correlation $0 \le {\bf x}_i {\bf
x}^T_j = \cos \theta \le 1$ as a similarity measure between two vectors,
and cluster vectors in $S$ accordingly.  However, when $90^o \le
|\theta| \le 180^o$, the correlation, ${\bf x}^T_i {\bf x}_j = \cos
\theta$, is a negative value.  The correlation is no more a good
measure for the geodesic distance.

\begin{figure}
\centering
\includegraphics[width=5cm]{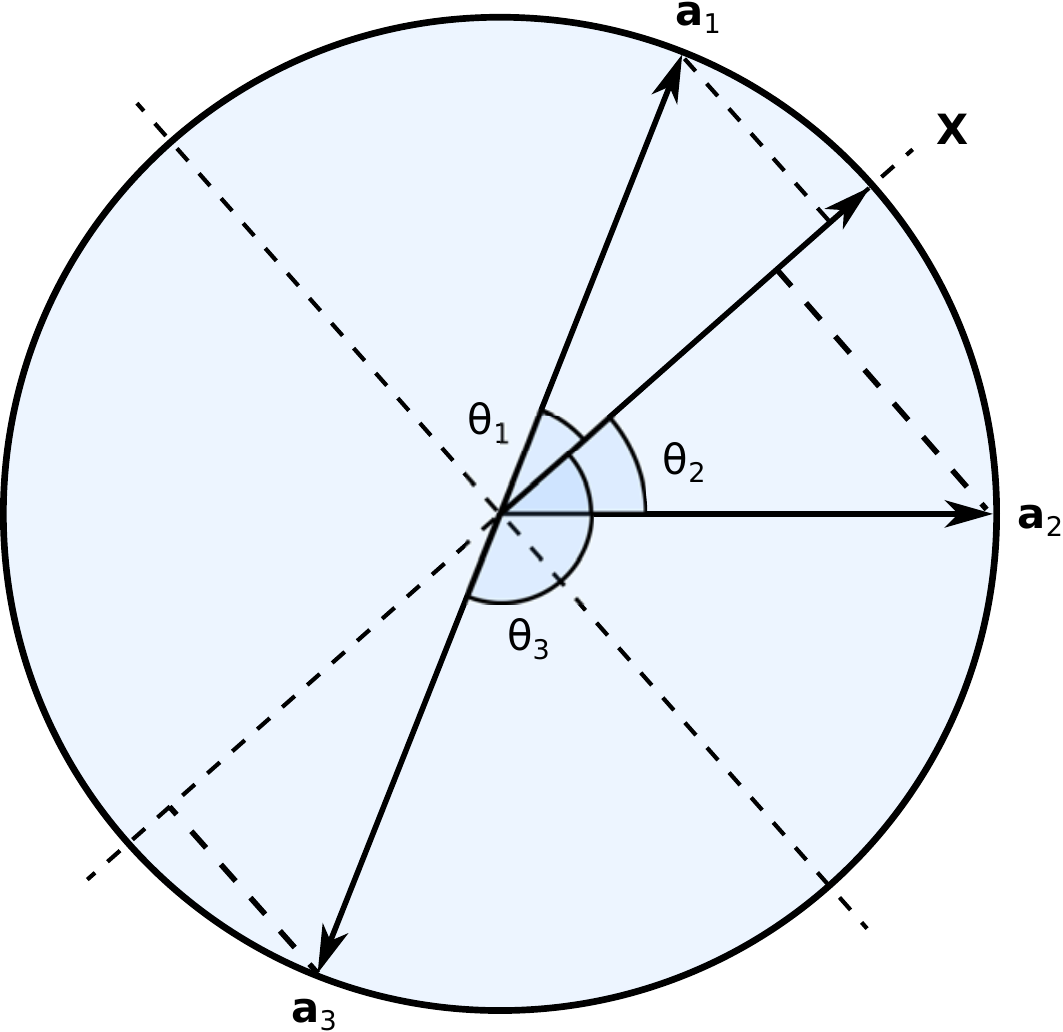}
\caption{An example to illustrate the need of correlation rectification
in the unit circle.}\label{fig:rectification}
\end{figure}

To show the necessity of rectification, a 2D example is illustrated in
Fig. \ref{fig:rectification}, where ${\bf x}$ and ${\bf a}_k$
($k=1,2,3$) denote an input and three anchor vectors on the unit circle,
respectively, and $\theta_i$ is their respective angle.  Since
$\theta_1$ and $\theta_2$ are less than 90 degrees, ${\bf a}^T_1 {\bf
x}$ and ${\bf a}^T_2 {\bf x}$ are positive. The correlation can be
viewed as a projection from an anchor vector to the input (and vice
versa). For positive correlations, the geodesic distance is a
monotonically decreasing function of the projection value.  The larger
the correlation, the shorter the distance. 

\begin{figure}
\centering
\includegraphics[width=3cm]{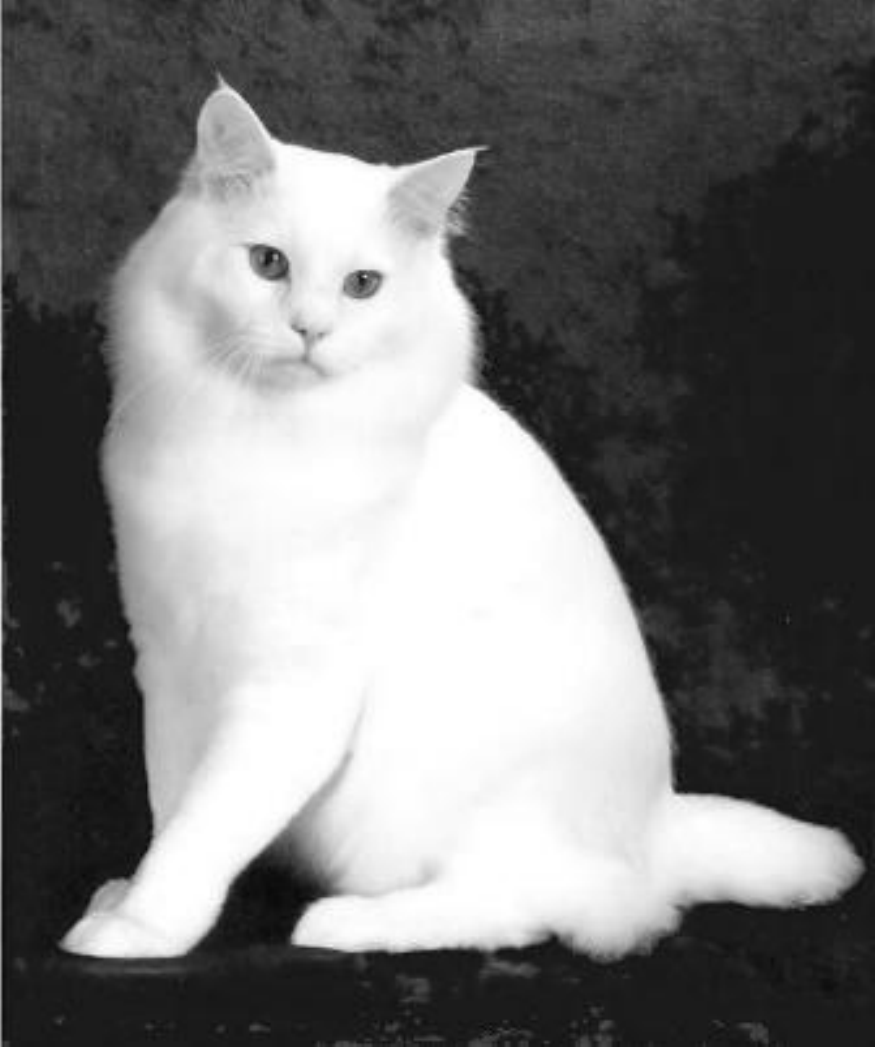} 
\includegraphics[width=3cm]{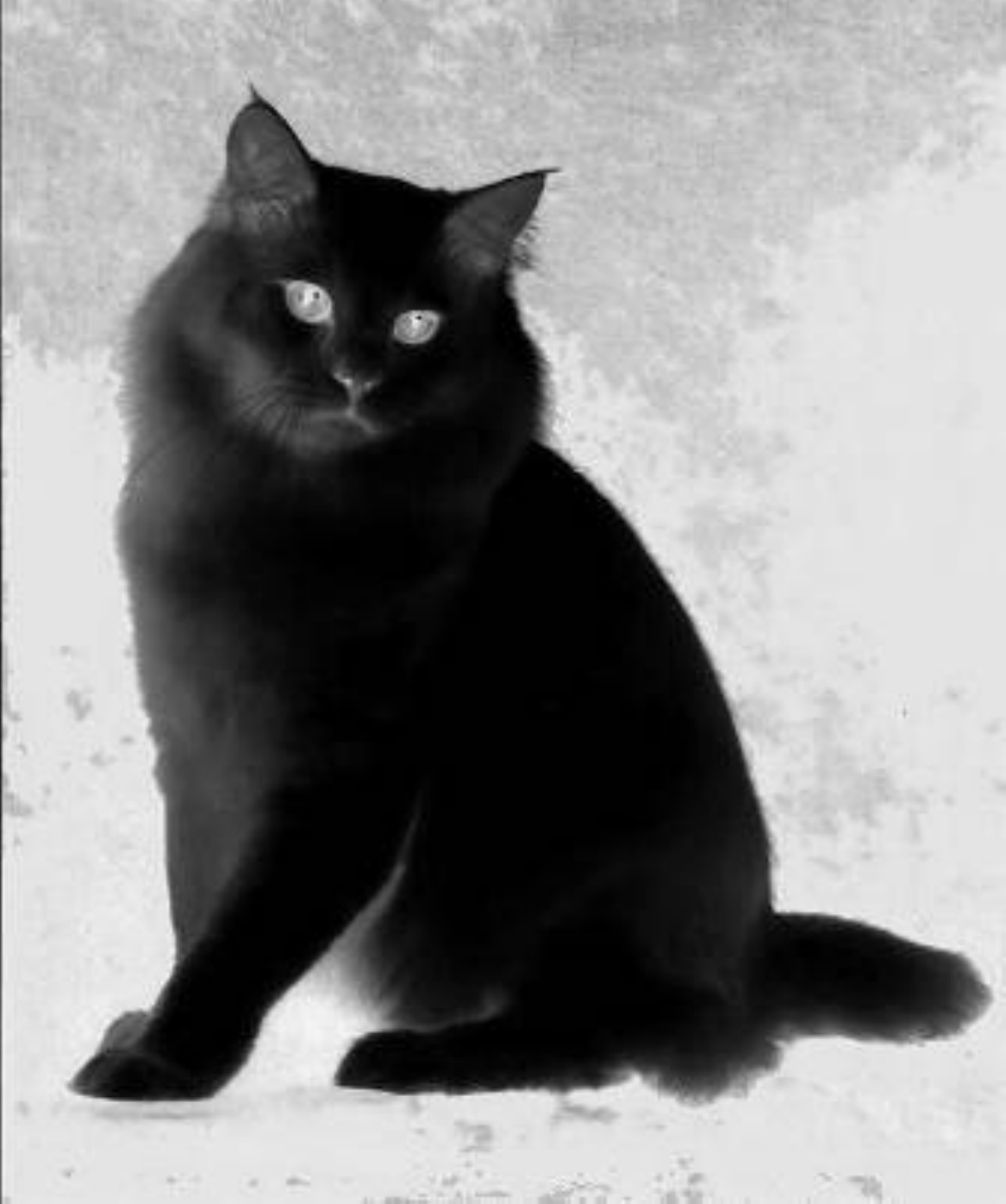}
\caption{A gray-scale cat image and its negative image. They are
negatively correlated after mean removal.  Their distance should be
large (a white cat versus a black cat).}\label{fig:cat}
\end{figure}

The angle, $\theta_3$, is larger than 90 degrees and correlation ${\bf
a}^T_3 {\bf x} $ is negative. The two vectors, ${\bf x}$ and ${\bf
a}_3$, are far apart in terms of the geodesic distance, yet their
correlation is strong (although a negative one). Consider the extreme
case. If ${\bf a}_3 =-{\bf x}$, ${\bf x}$ and ${\bf a}_3$ have the
farthest geodesic distance on the unit circle, yet they are fully
correlated but in the negative sense (see the example in Fig.
\ref{fig:cat}). For this reason, when the correlation value is negative,
it does not serve as a good indicator of the geodesic distance. 

One may argue that the negative sign could be used to signal a farther
geodesic distance. This is however not the case in a multi-layer RECOS
system if it does not have the nonlinear clipping operation. When two
RECOS units are in cascade, the filter weights of the 2nd RECOS unit can
take either positive or negative values. If the response of the 1st
RECOS unit is negative, the product of a negative response and a
negative filter weight will produce a positive value.  Yet, the product
of a positive response and a positive filter weight will also produce a
positive value. As a result, the system cannot differentiate these two
cases. Similarly, a system without rectification cannot differentiate
the following two cases: 1) a positive response at the first layer
followed by a negative filter weight at the second layer; and 2) a
negative response at the first layer followed by a positive filter
weight at the second layer. For this reason, it is essential to set a
negative correlation value (i.e. the response) at each layer to zero (or
almost zero) to avoid confusion in a multi-layer RECOS system. 

\begin{figure}
\centering
\includegraphics[width=10cm]{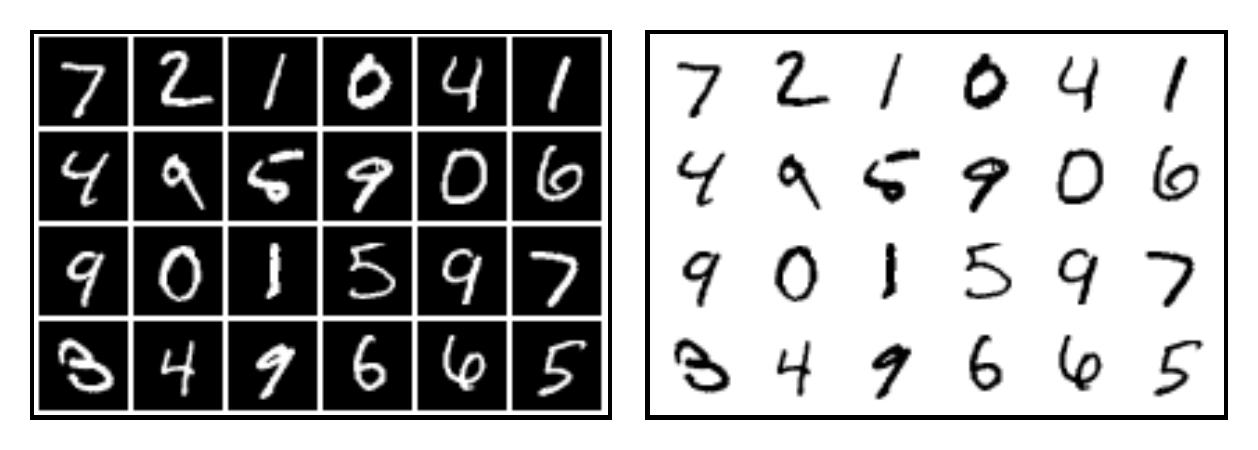}
\caption{Samples from the MNIST dataset: the orignal one (left) and the
gray-scale-reversed one (right).}\label{fig:mnist}
\end{figure}

We conducted an experiment to verify the importance of rectification. We
trained the LeNet-5 using the MNIST training dataset, and obtained a
correct recognition rate of 98.94\% for the MNIST test dataset. Then, we
applied the same network to gray-scale-reversed test images as shown in
Fig. \ref{fig:mnist}. The accuracy drops to 37.36\%. Next, we changed
all filter weights in C1 to their negative values while keeping the rest
of the network the same. The slightly modified LeNet-5 gives a correct
recognition rate of 98.94\% for the gray-scale-reversed test dataset but
37.36\% for the original test dataset. We can design a new network to
provide a high recognition rate to both test data in Fig.
\ref{fig:mnist} by doubling the number of anchor vectors in the first
layer. 

The above discussion can be written formally below.  Consider the case
where there are $K$ anchor vectors in the $N$-dimensional unit sphere,
denoted by ${\bf a}_k \in R^{N}$, $k=1, \cdots, K$.  For given ${\bf
x}$, its $K$ rectified correlations with ${\bf a}_k$, $k=1, \cdots, K$,
defines a nonlinear transformation from ${\bf x}$ to an output vector
\begin{equation}\label{eq:output0}
{\bf y}=(y_1, \cdots, y_k, \cdots, y_K)^T,
\end{equation}
where
\begin{equation}\label{eq:output}
y_k ({\bf x}, {\bf a}_k)= \max(0, {\bf a}^T_k {\bf x} ) 
\equiv {\rm Rec} ( {\bf a}^T_k {\bf x} ).
\end{equation}
The form in Eq. (\ref{eq:output}) is ReLU.  Other variants such as the
sigmoid function and the leaky ReLU are acceptable. As long as the
negative correlation values remain to be small, these vectors are weakly
correlated and they will not have a major impact on the final result. 

We can further generalize the RECOS model to a translated unit sphere
\small
\begin{equation}\label{eq:tunitsphere}
S_{\mu}=\left\{ {\bf x} \biggr\rvert ||{\bf x} - \mu {\bf 1} ||= 
\left[\sum_{n=1}^N (x_n- \mu)^2\right]^{1/2}=1 \right\}.
\end{equation}
\normalsize
where $\mu=\frac{1}{N} \sum_{n=1}^N x_n$ is the mean of all $x_n$'s and
${\bf 1}=(1, \cdots, 1, \cdots 1)^T \in R^N$ is a constant vector with
all elements equal to one. Sphere $S_{\mu}$ is a translation of $S$ with
a new center at $\mu {\bf 1}^T$. This generalization is desired for the
following reason. 

For vision problems, elements $x_n$, $n=1, \cdots, N$, of ${\bf x}$
denote $N$ pixel values of an input image, and $\mu$ is the mean of all
pixels.  If the input is a full image, its mean is the global mean that
has no impact on image understanding. It can be removed before the
processing. Thus, we set $\mu=0$. However, if an input image is large,
we often partition it into smaller patches, and process all patches in
parallel. In this case, the mean of each patch is a local mean.  It
should not be removed since an integration of local means provides a
coarse view of the full image. This corresponds to the general case in
Eq. (\ref{eq:tunitsphere}). 

Based on Eq. (\ref{eq:output}), the output with respect 
to $S_{\mu}$ can be written as ${\bf y}=(y_1, \cdots, y_K)$, where
\begin{equation}\label{eq:output2}
y_k ({\bf x} - \mu {\bf 1}, {\bf a}_k)= 
{\rm Rec} ( {\bf a}^T_k{\bf x} + \mu a_{k,0} ),
\end{equation}
and where $a_{k,0} = - \sum_{n=1}^N a_{k,n}$. By augmenting ${\bf x}$ 
and ${\bf a}_k$, with one additional element
\begin{equation}\label{eq:argment}
{\bf x'}=(\mu, x_1, \cdots, x_N)^T, \quad {\bf a'}_k = (a_{k,0}, 
a_{k,1}, \cdots, a_{k,N})^T,
\end{equation}
we can re-write Eq. (\ref{eq:output2}) as
\begin{equation}\label{eq:output3}
y_k ({\bf x'}, {\bf a'}_k)= {\rm Rec} ( {\bf a'}^T_k {\bf x'}), \quad k=1, \cdots, K.
\end{equation}
Although ${\bf x'}, {\bf a'}_k \in R^{(N+1)}$, they only have $N$
independent elements since their first elements are computed from the
remaining $N$ elements. 

Furthermore, the length of the input and anchor vectors may not be
one.  We use ${\bf x''}$ and ${\bf a''}_k$ to denote the general case. Then,
we can set
\begin{equation}\label{eq:normalization}
{\bf x'} \equiv \frac{\bf x''}{||{\bf x''}||}, \quad 
{\bf a'}_k \equiv \frac{{\bf a''}_k}{||{\bf a''}_k||}.
\end{equation}
Then, Eq. (\ref{eq:output3}) can be re-written as
\begin{equation}\label{eq:output4}
y_k ({\bf x''}, {\bf a''}_k)= ||{\bf x''}|| ||{\bf a''}_k|| 
{\rm Rec} ( {\bf a'}^T_k {\bf x'} ).
\end{equation}

If there are $K$ frequently occuring patterns in input data to form $K$
clusters, we can assign anchor vector, ${\bf a}_k$, to the centroid of
the $k$th cluster, $k=1, \cdots,K$. Then, data points in this cluster
will generate a strong correlation with ${\bf a}_k$.  They will generate
weaker correlations with other anchor vectors. Along this line, it is
worthwhile to mention that it was observed in \cite{AISTATS2011} that
the K-means clustering is effective in a single-layer network.  A CNN
consists of multiple RECOS units working cooperatively, and they can be
organized into multiple layers. The advantage of multi-layer CNNs will
be explained in the next section. 

As discussed earlier, we can reverse the sign of all filter weights in
the first layer (i.e. C1) while keeping the rest of the LeNet-5 the same
to obtain a modified LeNet-5. Then, the modified LeNet-5 has the same
recognition performance against the gray-scale-reversed test dataset.
This observation can actually be proved mathematically. The input of a
gray-scale-reversed image to the first layer can be written as
\begin{equation}\label{eq:x_r}
{\bf x}_r=255 {\bf 1} - {\bf x}, 
\end{equation}
where ${\bf x}$ is the input from the original image.  The mean of the
elements in ${\bf x}_r$, denoted by $\mu_r$, is equal to
$\mu_r=255-\mu$, where $\mu$ is the mean of the elements in ${\bf x}$.
Furthermore, the anchor vectors become
\begin{equation}\label{eq:a_r}
{\bf a}_{r,k} = - {\bf a}_{k},
\end{equation}
where ${\bf a}_k$ is the anchor vector of the LeNet-5. Then, by
following Eq.  (\ref{eq:output2}), the output from the first layer of
the modified LeNet-5 against the gray-scale-reversed image can be
written as
\begin{eqnarray}\label{eq:output2_r}
y_k ({\bf x}_r - \mu_r {\bf 1}, {\bf a}_{r,k}) & = &
y_k (255 {\bf 1} - {\bf x} - (255-\mu) {\bf 1}, -{\bf a}_{k}) \\
& = & y_k ({\bf x} - \mu {\bf 1}, {\bf a}_{k}), \label{eq:output2_rr}
\end{eqnarray}
where the last term is the output from the first layer of the LeNet-5
against the original input image. In other words, the two systems
provide identical output vectors to be used in future layers. 

\section{Advantages of Cascaded Layers?}

\begin{figure}
\centering
\includegraphics[width=5cm]{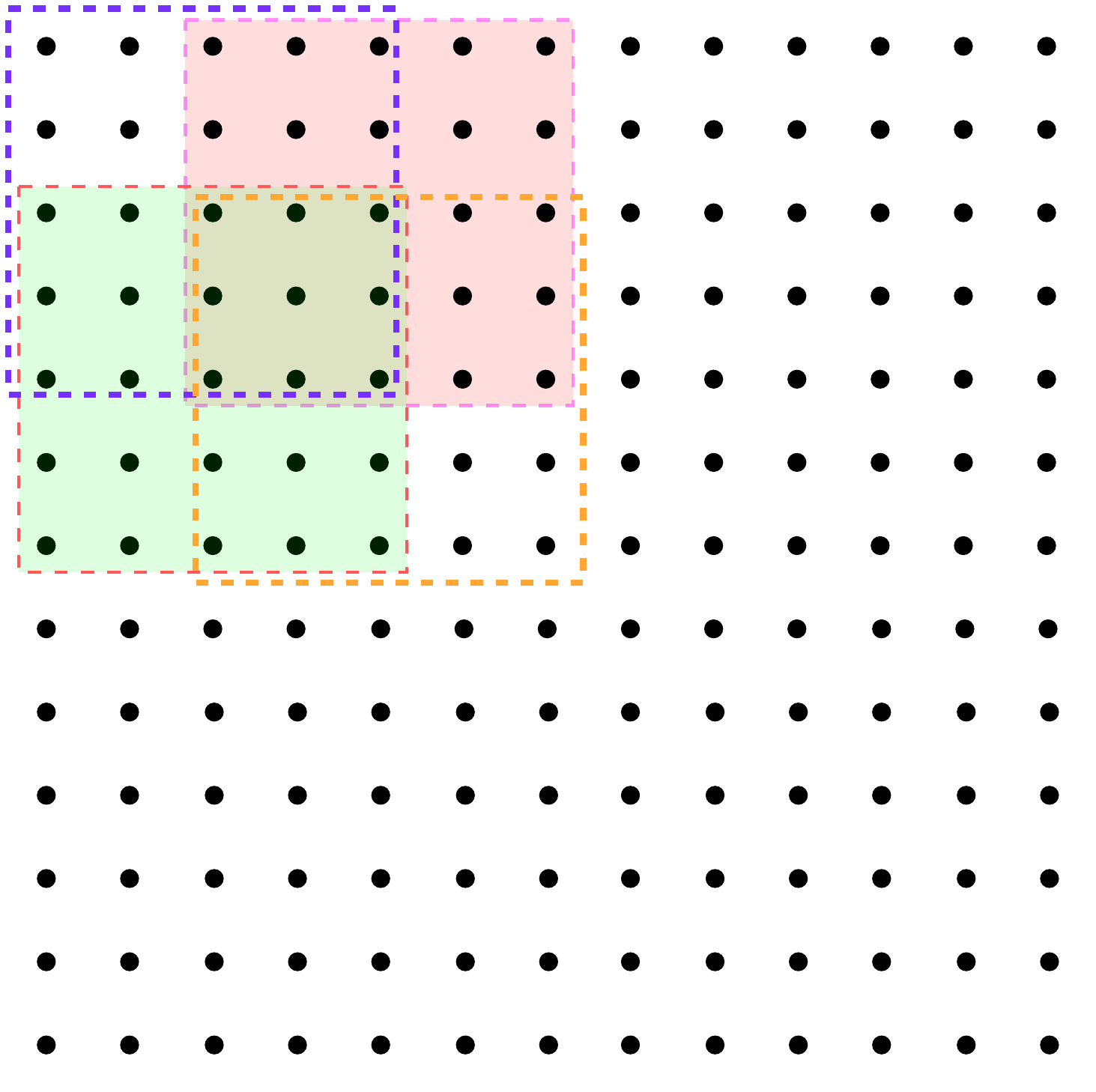}
\caption{The receptive fields of the first- and the second-layer filters
of the LeNet-5, where each dot denotes a pixel in the input image, the
$5 \times 5$ window indicates the receptive field of the first-layer
filter and the whole $13\times 13$ window indicates the receptive field
of the second-layer filter. The second-layer filter accepts the outputs
from $5 \times 5=25$ first-layer filters. Only four of them are shown in
the figure for simplicity.}\label{fig:receptive}
\end{figure}

The LeNet-5 is essentially a neural network with two convolutional
layers since the compound operations of convolution/sampling/nonlinear
clipping are viewed as one complete layer in the modern neural network
literature. The input to the first layer of the LeNet-5 is a purely
spatial signal while the input to the second layer of the LeNet-5 is a
hybrid spectral-spatial signal consisting of spatial signals from 6
spectral bands. The spatial coverage of a filter with respect to the
source image is called its receptive field. The receptive fields of the
first and the second layers of the LeNet-5 are $5\times5$ and $13 \times
13$, respectively. For each spatial location in the $13 \times 13$
receptive field, it may be covered by one, two or four layer-one filters
as shown in Fig.  \ref{fig:receptive}. In the following, we conduct a
mathematical analysis on the behavior of the cascaded systems.  This
analysis sheds light on the advantage of deeper networks.  In the
following discussion, we begin with the cascade of one layer-1 RECOS
unit and one layer-2 RECOS unit, and then generalize it to the cascade of
multiple layer-1 RECOS units to one layer-2 RECOS unit. For simplicity,
the means of all inputs are assumed to be zero. 

\begin{figure}
\centering
\includegraphics[width=6cm]{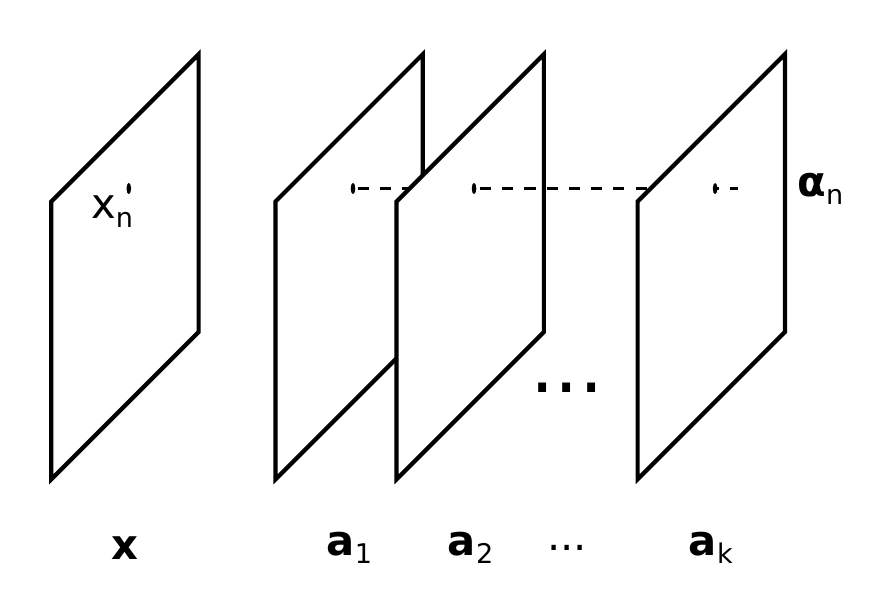}
\caption{Visualization of anchor-position vector ${\boldsymbol \alpha}_n$.}
\label{fig:alpha}
\end{figure}

{\bf One-to-One Cascade.} We define two anchor matrices:
\begin{equation}\label{eq:anchor_matrix}
{\bf A} = \left[ {\bf a}_1, \cdots,  {\bf a}_k  \cdots, {\bf a}_K \right], \;
{\bf B} = \left[ {\bf b}_1, \cdots,  {\bf b}_l  \cdots, {\bf b}_L \right], 
\end{equation}
whose column are anchor vectors ${\bf a}_k$ and ${\bf b}_l$ of the two
individual RECOS units. Clearly, ${\bf A} \in R^{N\times K}$ and ${\bf
B} \in R^{K \times L}$. To make the analysis tractable, we begin with
the correlation analysis and will take the nonlinear rectification
effect into account at the end.  For the correlation part, let ${\bf y}
= {\bf A}^T {\bf x}$ and ${\bf z} = {\bf B}^T {\bf y}$. Then, we have
\begin{equation}\label{eq:C}
{\bf z} =  {\bf B}^T {\bf A}^T {\bf x} = {\bf C}^T{\bf x} , \; 
{\bf C} \equiv {\bf A} {\bf B}. 
\end{equation}
Clearly, ${\bf C} \in R^{N \times L}$ with its $(n,l)$th element equal to
\begin{equation}\label{eq:compound}
c_{n,l} = {\boldsymbol \alpha}^T_n {\bf b}_l,
\end{equation}
where ${\boldsymbol \alpha}^T_n \in R^K$ is the $n$th row vector of {\bf
A}.  The meaning of ${\boldsymbol \alpha}_n$ can be visualized in Fig.
\ref{fig:alpha}.  Mathematically, we decompose
\begin{equation}\label{x-decompose}
{\bf x} = \sum_{n=1}^N x_n {\bf e}_n,
\end{equation}
where ${\bf e}_n \in R^N$ is the $n$th coordinate-unit-vector. Then,
\begin{equation}\label{alpha_n}
{\boldsymbol \alpha}_n={\bf A}^T {\bf e}_n.
\end{equation}
Since ${\boldsymbol \alpha}_n$ captures the position information of
anchor vectors in ${\bf A}$, it is called the anchor-position vector.
Finally, we apply the rectification to all negative elements in ${\bf C}$
to obtain an anchor matrix ${\bf C'}$ from ${\bf x}$ to ${\bf z'}$:
\begin{equation}\label{eq:r-compound}
{\bf z'} = {\bf C'}^T {\bf x}, \quad {\bf C'} = [c'_{n,l}]_{N \times L}, 
\end{equation}
where 
\begin{equation}\label{eq:compound}
c'_{n,l} = {\rm Rec}(c_{n,l}) = {\rm Rec}({\boldsymbol \alpha}^T_n {\bf b}_l).
\end{equation}

Rigorously speaking, ${\bf z}$ and ${\bf z'}$ are not the same. The
former has no rectification operation while the latter applies the
rectification operation to the matrix product. Since the actual system
applies the rectification operations to the output of both layers and
its final result, denoted by ${\bf z''}$, can be different from ${\bf
z}$ and ${\bf z'}$.  Here, we are most interested in areas where ${\bf
z} \approx {\bf z'} \approx {\bf z''}$ in the sense that they go through
the same rectification processes in both layers. Under this assumption,
our above analysis holds for the unrectified part of the input. 

{\bf Many-to-One Cascade}. It is straightforward to generalize
one-to-one cascaded case to the many-to-one cascaded case.  The
correlation of the first-layer RECOS units can be written as
\begin{equation}\label{eq:XY1}
{\bf Y} = {\bf A}^T {\bf X} ,
\end{equation}
where
\begin{equation}\label{eq:XY2}
{\bf Y} =\left[ {\bf y}_1, \cdots {\bf y}_P \right], \;
{\bf X} =\left[ {\bf x}_1, \cdots {\bf x}_P \right], \;
\end{equation}
There are $P$ RECOS units working in parallel in the first layer. They
cover spatially adjacent regions yet share one common anchor matrix.
They are used to extract common representative patterns in different
regions. The correlation of the second-layer RECOS can be expressed as
\begin{equation}\label{eq:YZ1}
{\bf z} = {\bf B}^T {\bf \tilde{y}},
\end{equation}
where ${\bf z} \in R^L$, ${\bf B} \in R^{PK \times L}$ and ${\bf
\tilde{y}}=({\bf y}^T_1 , \cdots, {\bf y}^T_P)^T \in R^{PK}$ is formed
by the cascade of $P$ output vectors of the first-layer RECOS units.

Anchor matrix {\bf A} extracts representative patterns in different
regions while anchor matrix {\bf B} is used to stitch these
spatially-dependent representative patterns to form larger
representative patterns.  For example, consider a large lawn composed by
multiple grass patches.  Suppose that the grass patterns can be captured
by an anchor vector in ${\bf A}$. Then, an anchor vector in {\bf B} will
provide a superposition rule to stitch these spatially distributed
anchor vectors of grass to form a larger lawn. 

{\bf Comparison of One-Layer and Two-Layer Systems.} To explain
the advantage of deeper networks, we compare the two-layer system 
in Eq. (\ref{eq:r-compound}) with the following one-layer system
\begin{equation}\label{eq:D}
{\bf z} = {\bf D}^T {\bf x},
\end{equation}
where ${\bf D}=[ {\bf d}_1, \cdots, {\bf d}_L ] \in R^{N \times L}$ is
an anchor matrix with ${\bf d}_l$ as its anchor vector.  Anchor matrices
${\bf A}$ and ${\bf D}$ essentially play the same role in capturing global
frequently occurring patterns in ${\bf x}$. However, the two-stage system 
has additional anchor matrix, ${\bf B}$, in capturing representative
patterns of ${\bf y}$. It is best to examine anchor vectors of ${\bf C'}$ to
understand the compound effect of ${\bf A}$ and ${\bf B}$ fully. 
Based on Eq.(\ref{eq:r-compound}), anchor vectors of ${\bf C'}$ are
\begin{equation}\label{eq:anchor-compound}
{\bf c'}_l = (c'_{1,l}, \cdots, c'_{n,l})^T, \; l=1, \cdots, L,
\end{equation}
where $c'_{n,l}$ is the rectified inner product of ${\boldsymbol
\alpha}_n$ and ${\bf b}_l$ as given in Eq. (\ref{eq:compound}).  Anchor
vectors ${\bf a}_k$ capture representative global patterns, but they are
weak in capturing position sensitive information. This shortcoming can
be compensated by modulating ${\bf b}_l$ with elements of
anchor-position vector ${\boldsymbol \alpha}_n$. 

\begin{figure}[!h]
\centering
\includegraphics[width=0.95\textwidth]{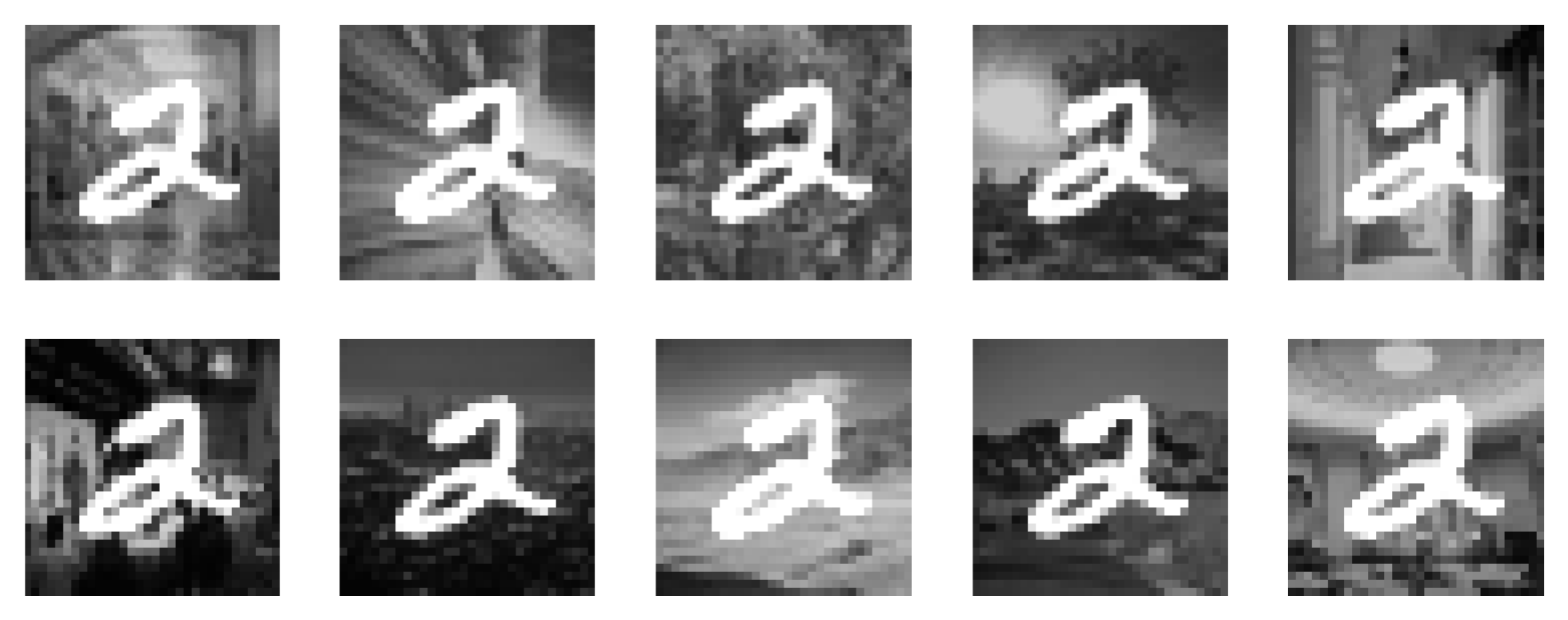} \\
\includegraphics[width=0.95\textwidth]{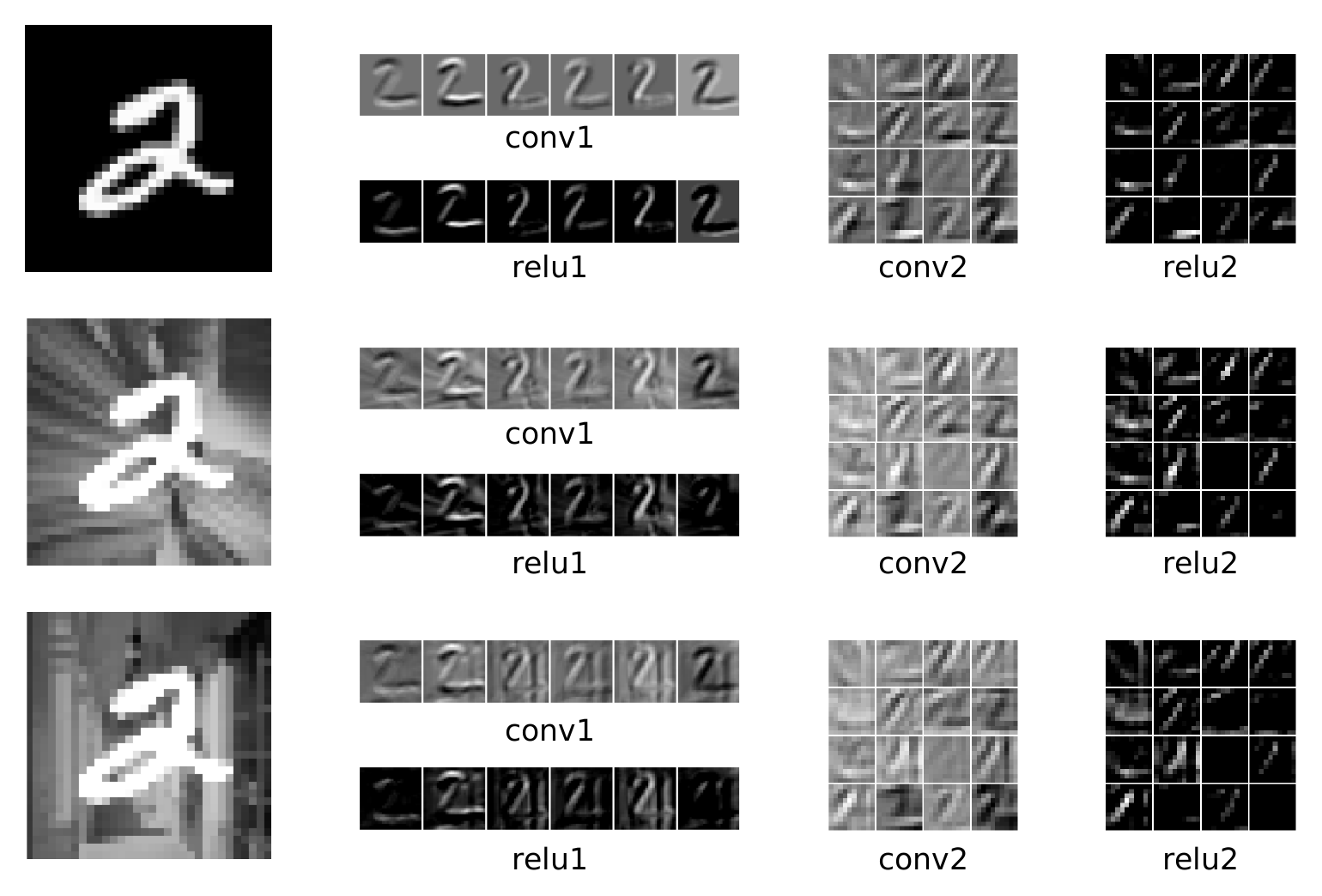} 
\caption{The MNIST dataset with 10 different background scenes are shown
in the top two rows while the output images in 6 spectral channels and
16 spectral channels of the first-layer and the second-layers with
respect to the input images given in the leftmost column are shown in
the bottom three rows. The structured background has an impact on the
6 channel responses at the first layer yet their impact on the 16
channel responses at the second layer diminishes. This phenomenon can
be explained by the analysis given in Section 3.}\label{fig:background}
\end{figure}

We use an example to illustrate this point. First, we modify the MNIST
training and testing datasets by adding ten different background scenes
randomly to the original handwritten digits in the MNIST dataset
\cite{LeNet1998}.  They are shown in the top two rows in Fig.
\ref{fig:background}. For the bottom three rows, we show three input
digital images in the leftmost column, the six spectral output images
from the convolutional layer and the ReLU layer in the middle column and
the 16 spectral output images in the right two columns.  It is difficult
to find a good anchor matrix of the first layer due to background
diversity. However, background scenes in these images are not consistent
in the spatial domain while foreground digits are.  As a result, they
can be filtered out more easily by modulating anchor vectors ${\bf b}_l$
in the second layer using the anchor-position vector, ${\boldsymbol
\alpha}_n$, in the first layer. 

Experiments are further conducted to provide supporting evidences.
First, we add one of the ten complex background scenes to test images
randomly and pass them to the LeNet-5 trained with the original MNIST
data of clean background. The recognition rate drops from 98.94\% to
90.65\%. This is because that this network has not yet seen any
handwritten digits with complex background. Next, we modify the MNIST
training data by adding one of the ten complex background scenes
randomly and train the LeNet-5 using the modified MNIST data again. The
newly trained network has a correct classification rate of 98.89\% and
98.86\% on the original and the modified MNIST test datasets,
respectively.  We see clearly that the addition of a sufficiently
diverse complex background scenes in the training data has little impact
on the capability of the LeNet-5 in recognizing images of clean
background.  This is because that the complex background is not
consistent with labeled digits and, as a result, the network can focus
on the foreground digits and ignore background scenes through the
cascaded two-layer architecture. Our previous analysis provides a
mathematical explanation to this experimental result. It is also
possible to understand this phenomenon by visualizing CNN filter
responses \cite{simonyan2013deep,zeiler2014visualizing,zhou2014object}.

\begin{figure}[!t]
\centering
\includegraphics[width=0.42\linewidth]{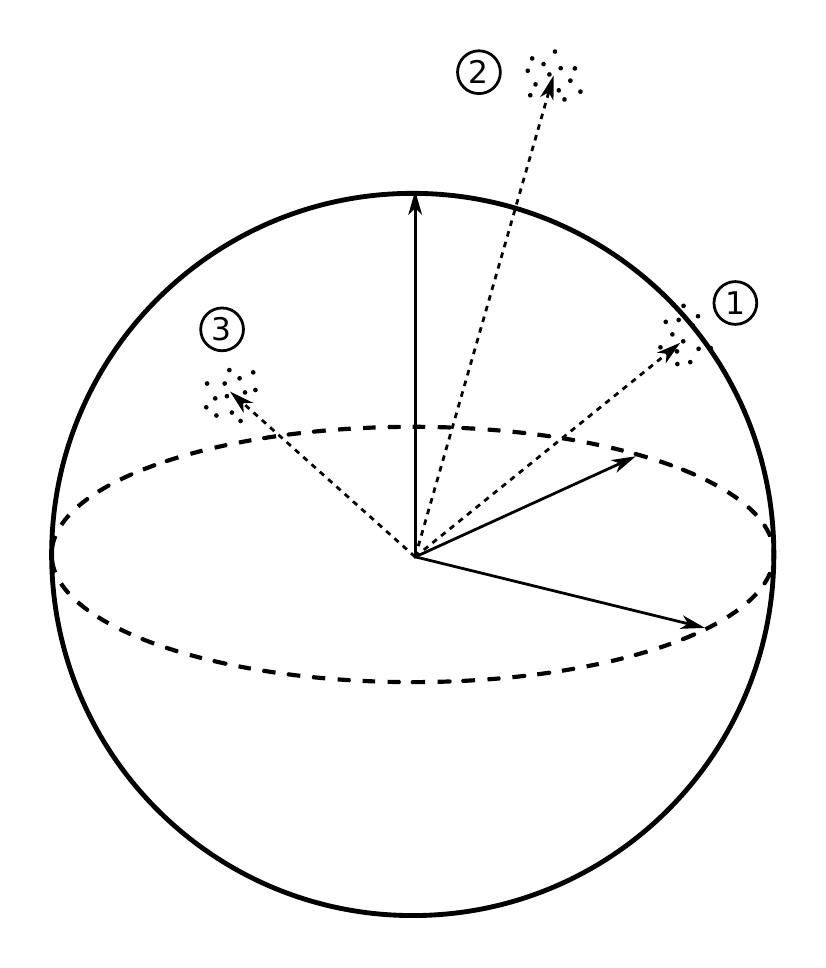} \hspace{.1in}
\includegraphics[width=0.42\linewidth]{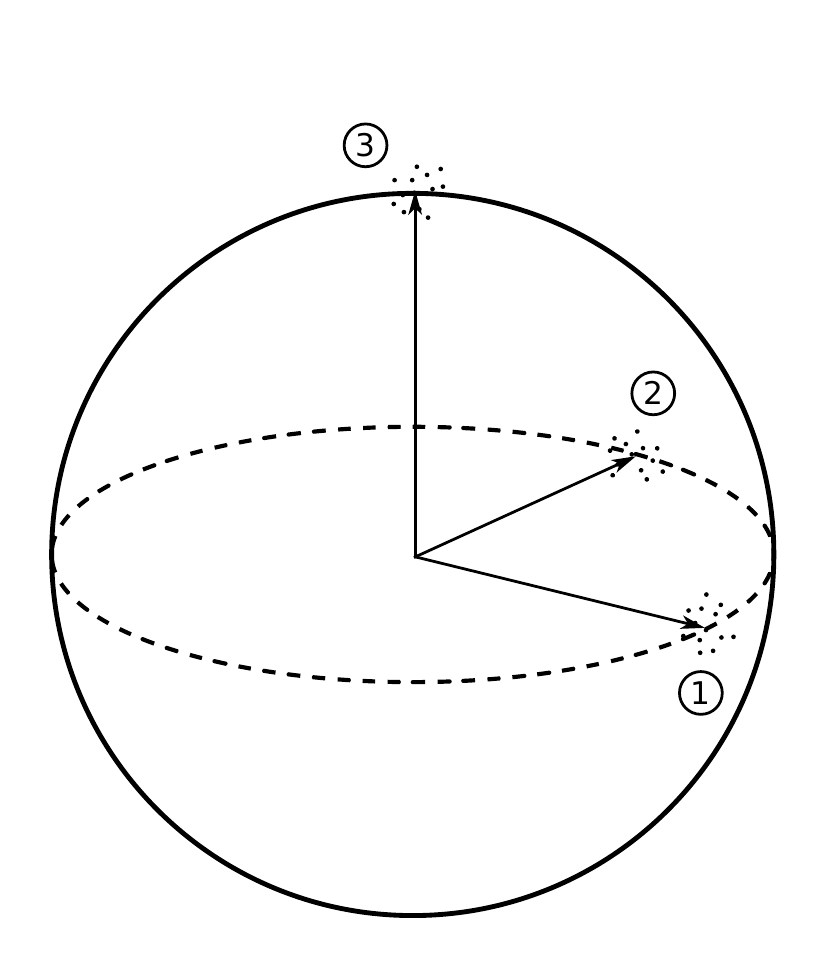}\\
(a) \hspace{1.5in}  (b)
\caption{Illustration of functions of (a) C5 and (b) F6.}\label{fig:aa}
\end{figure}

{\bf Role of Fully Connected Layers.} A CNN can be decomposed into two
subnetworks (or subnets): the feature extraction subnet and the decision
subnet. For the LeNet-5, the feature extraction subnet consists of C1,
S2, C3 and S4 while the decision subnet consists of C5, F6 and Output as
shown in Fig. \ref{fig:aa}. The decision subnet has the following three
roles: 1) converting the spectral-spatial feature map from the output of
S4 into one feature vector of dimension 120 in C5; 2) adjusting anchor
vectors so that they are aligned with the coordinate-unit-vector with
correct feature/digit pairing in F6; and 3) making the final digit
classification decision in Output. 

The functions of C5 and F6 are illustrated in Fig. \ref{fig:aa}(a) and
(b), respectively.  C5 assigns anchor vectors to feature clusters as
shown in Fig.  \ref{fig:aa}(a).  There are 120 anchor vectors in the
400-D space to be assigned (or trained) in the LeNet-5.  Then, an anchor
matrix is used in F6, to rotate and rescale anchor vectors in C5 to
their new positions.  The objective is to ensure the feature cluster of
an object class to be aligned with the coordinate-unit-vector of the
same object class for decision making in the Output.  This is
illustrated in Fig.  \ref{fig:aa} (b). Every coordinate-unit-vector in
the Output is an anchor vector, and each of them represents a digit
class. The softmax rule is widely used in the Output for final decision
making.

{\bf Multi-Layer CNN.} We list the traditional layer names, RECOS
notations, their input and the output vector dimensions of the LeNet-5
in Table \ref{table:LeNet}.  The output vector dimension is the same as
the number of anchor vectors in the same layer.  Vector augmentation is
needed in $S^1$ since their local mean could be non-zero.  However, it
is not needed in $S^2$, $S^3$ $S^4$ and $S^5$ since the global mean is
removed. 

\begin{table} [!th]
\centering
\caption{The specification of RCS units used in the LeNet-5, where the
third column ($N$) shows the dimension of the input and the fourth column 
($K$) shows the dimension of the output of the corresponding layer. Note
that $K$ is also the number of anchor vectors.}\label{table:LeNet}
\begin{tabular}{|c|l|c|c|} \hline
LeNet-5 & RECOS & N & K \\ \hline \hline
C1/S2 & $S^1$ & $(5 \times 5)+1$ & $6$  \\
C3/S4 & $S^2$ & $(6 \times 5 \times 5)$ & $16$  \\
C5 & $S^3$ & $16 \times 5 \times 5$ & $120$  \\
F6 & $S^4$ & $120 \times 1 \times 1$ & $84$  \\
Output & $S^5$ & $84 \times 1 \times 1$ & $10$ \\ \hline
\end{tabular}
\end{table}

\begin{figure} [!th]
\centering
\includegraphics[width=10cm]{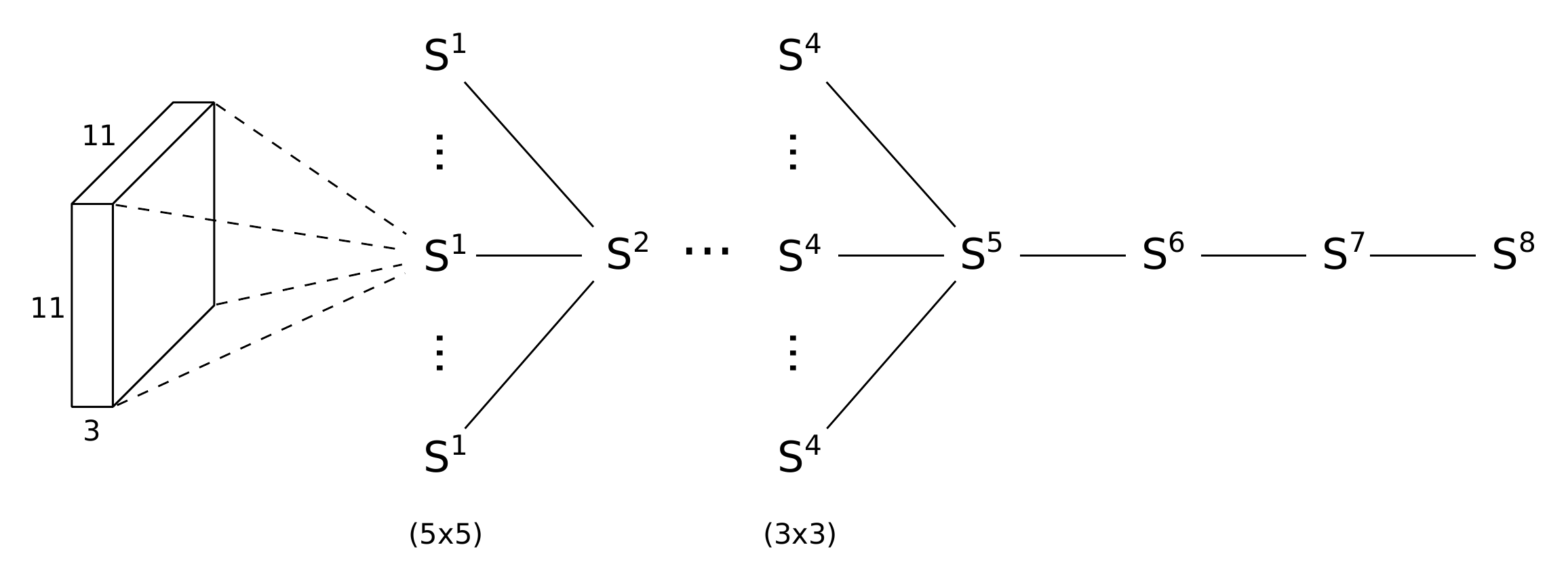}
\caption{The organization of the AlexNet using tree-structured RECOS 
units.}\label{fig:tree}
\end{figure}

\begin{table} [!th]
\centering
\caption{The specification of RCS units used in the AlexNet, where the
third column ($N$) shows the dimension of the input and the fourth column 
($K$) shows the dimension of the output of $S^l$, where $l=1,\cdots,8$.
$K$ is also the number of anchor vectors.}\label{table:alexnet}
\begin{tabular}{|c|l|c|c|} \hline
AlexNet & RECOS & N & K \\ \hline \hline
$Conv\_1$ & $S^1$ & $(3 \times 11 \times 11)+1$ & $96$  \\
$Conv\_2$ & $S^2$ & $(96 \times 5 \times 5)+1$ & $256$  \\
$Conv\_3$ & $S^3$ & $(256 \times 3 \times 3)+1$ & $384$  \\
$Conv\_4$ & $S^4$ & $(384 \times 3 \times 3)+1$ & $384$  \\
$Conv\_5$ & $S^5$ & $(384 \times 3 \times 3)+1$ & $256$ \\
$FC\_6$   & $S^6$ & $256 \times 1 \times 1$ & $4096$ \\
$FC\_7$   & $S^7$ & $4096 \times 1 \times 1$ & $4096$ \\
$FC\_8$    & $S^8$ & $4096 \times 1 \times 1$ & $1000$ \\ \hline
\end{tabular}
\end{table}

Next, we use the AlexNet proposed by Krizhevsky {\em et al.}
\cite{NIPS2012_AlexNet} as another example to illustrate a multi-layer
CNN. Its RECOS representation is shown in Fig. \ref{fig:tree} and Table
\ref{table:alexnet}.  We denote the RECOS units at the $l$th level by
$S^l$ ($l=1, \cdots 8$).  The input to $S^1$ is a color image patch of
size $11\times 11$ with $R$, $G$ and $B$ channels.  The covered regions
become larger as we proceed from $S^1$ to $S^5$.  They are used to
capture representative visual patterns of various sizes and at different
spatial locations. 

\section{Conclusion and Open Problems}

In this work, a RECOS model was adopted to explain the role of the
nonlinear clipping function in CNNs, and a simple matrix analysis was
used to explain the advantage of the two-layer RECOS model over the
single-layer RECOS model.  The proposed RECOS mathematical model is
centered on the selection of anchor vectors.  CNNs do offer a very
powerful tool for image processing and understanding.  There are however
a few open problems remaining in CNN interpretability and wider
applications. Some of them are listed below. 

\begin{enumerate}
\item {\bf Application-Specific CNN Architecture} \\
In CNN training, the CNN architecture (including the layer number and
the filter number at each layer, etc.) has to be specified in advance.
Given a fixed architecture, the filter weights are optimized by an
end-to-end optimization framework. Generally speaking, simple tasks can
be well handled by smaller CNNs. However, there is still no clear
guideline in the CNN architecture design for a class of applications.
The anchor-vector viewpoint encourages us to examine the properties of
source data carefully. A good understanding of source data distribution
contributes to the design of more efficient CNN architectures and more
effective training. 
\item {\bf Robustness to Input Variations} \\
The LeNet-5 was shown to be robust with respect to a wide range of
variations in \cite{LeNet1998}.  Yet, the robustness of CNNs is
challenged by recent studies, e.g.  \cite{SzegedyZSBEGF13}. It is an
interesting topic to understand the causes of these problems so as to
design more error-resilient CNNs. 
\item {\bf Weakly Supervised Learning} \\
The training of CNNs demand a large amount of labeled data. It is
expensive to collect labeled data. Furthermore, the labeling rules could
be different from one dataset from another even for the same
applications. It is important to reduce the labeling burden and allow
CNN training using partially and flexibly labeled data. In other words,
we need to move from the heavily supervised learning to weakly
supervised learning to make CNNs widely applicable.
\item {\bf Effective Back-propagation Training} \\
Effective back-propagation training is important as CNNs become more and
more complicated nowadays.  Several back-propagation speed-up schemes
have been proposed. One is dropout \cite{NIPS2012_AlexNet}.  Another one
is to inject carefully chosen noise to achieve faster convergence as
presented in \cite{audhkhasi2016noise}. New methods along this direction
are very much in need. 
\end{enumerate}

\section*{Acknowledgment}

The author would like to thank Mr. Zhehang Ding's help in running
experiments and drawing figures for this article.  This material is
based on research sponsored by DARPA and Air Force Research Laboratory
(AFRL) under agreement number FA8750-16-2-0173. The U.S. Government is
authorized to reproduce and distribute reprints for Governmental
purposes notwithstanding any copyright notation thereon. The views and
conclusions contained herein are those of the authors and should not be
interpreted as necessarily representing the official policies or
endorsements, either expressed or implied, of DARPA and Air Force
Research Laboratory (AFRL) or the U.S. Government. 

\section*{\refname}
\bibliographystyle{elsarticle-num} 
\bibliography{CNN}

\end{document}